\documentclass[conference, letterpaper]{IEEEtran}

\ifCLASSINFOpdf

\else

\fi

\ifCLASSINFOpdf
   \usepackage[pdftex]{graphicx}
\else
\fi

\usepackage[cmex10]{amsmath}
\usepackage{color}
\usepackage{cite}
\usepackage{fancyhdr}
\usepackage[caption=false,font=footnotesize]{subfig}
\usepackage{enumitem}
\usepackage{hyperref}
\renewcommand{\thispagestyle}[2]{}

\fancypagestyle{plain}{
        \fancyhead{}
        \fancyhead[C]{first page center header}
        \fancyfoot{}
        \fancyfoot[C]{first page center footer}
}
\begin{document}

%
\title{AN INTEGRATED NPL APPROACH TO SENTIMENT ANALYSIS IN SATISFACTION SURVEYS}

\author{\IEEEauthorblockN{\textbf{Pinto Luque Edson Bladimir}}
\IEEEauthorblockA{Faculty of Statistic and Computer Engineering\\Universidad Nacional del Altiplano de Puno, P.O. Box 291\\
Puno - Perú \\
Email: epintol@est.unap.edu.pe}}


%


\maketitle

\begin{abstract}
The research project aims to apply an integrated approach to natural language processing (NLP) to satisfaction surveys. It will focus on understanding and extracting relevant information from survey responses, analyzing feelings, and identifying recurring word patterns. NLP techniques will be used to determine emotional polarity, classify responses into positive, negative, or neutral categories, and use opinion mining to highlight participants' opinions. This approach will help identify the most relevant aspects for participants and understand their opinions in relation to those specific aspects.

A key component of the research project will be the analysis of word patterns in satisfaction survey responses using NPL. This analysis will provide a deeper understanding of feelings, opinions, and themes and trends present in respondents' responses. The results obtained from this approach can be used to identify areas for improvement, understand respondents' preferences, and make strategic decisions based on analysis to improve respondent satisfaction.
\end{abstract}


\begin{IEEEkeywords}
Sentimental Analysis; BERT; NPL
\end{IEEEkeywords}

%
\IEEEpeerreviewmaketitle

\section{INTRODUCTION}
Our research focuses on analyzing satisfaction survey responses for success and sustainability, as understanding customer sentiment and opinions is crucial to identifying areas for improvement and making informed decisions.

Opinion mining, commonly known as sentiment analysis and emotion recognition\cite{Gandhi2023} \cite{Xu2021} is an increasingly popular NLP technique in this developing area of study is to enable intelligent systems to observe, infer and understand human emotions. \cite{Denecke2023} Computer science, psychology, social sciences and cognitive sciences are included in the interdisciplinary discipline.\cite{Gandhi2023}. These techniques have proven effective in fields such as product evaluation, social network analysis\cite{Aldinata2023} \cite{Carlos2018}, and business decision making.\cite{Liu2012} This technique is used to collect and examine popular sentiment and ideas.\cite{Gandhi2023} therefore by applying NLP techniques in satisfaction surveys, organizations can gain deeper insights into respondents' preferences, needs and concerns, giving them a competitive advantage and enabling informed decisions, learning, communication and situational awareness in human-centric environments..\cite{Bhattacharjee2022} \cite{Shaik2023} \cite{Gandhi2023} This valuable insight helps organizations adapt their products and services, improve their strategies, and adapt to different situations.

Sentiment analysis is a system that automatically detects the opinions expressed in the comments \cite{Benchimol2022}.Sentiment analysis focused primarily on sentiment at the text and sentence level. It is challenging to discover numerous sentiment features present in the text, as these sentiment analyzes often only consider the signal element of the sentiment. However, human emotions are complicated.\cite{Denecke2023}

Sentiment analysis research can be categorized into the following three subfields based on granularity: aspect-level sentiment analysis, sentence-level sentiment analysis, and document-level sentiment analysis.\cite{Jiang2023}
Determining the polarity of the sentiment of a sentence specifically in the suggestions and opinions that the user provides when answering a satisfaction survey is what we intend to do in this article. Polarity typically includes positive, negative, and neutral.\cite{Chen2022} All of this is done to enhance the first responder experience and provide an accurate and comprehensive view of perceptions and requirements, supporting informed decision making.\cite{Umair2022}

\section{METHODOLOGY}
Natural language processing (NLP) is a branch of linguistics and artificial intelligence that focuses on how human and machine language interact.\cite{Rakshitha2021} focuses on methods for managing or analyzing massive amounts of data that produce recommendations for the workshop course. as a result, the computer being able to "understand" the document's concept.\cite{Denecke2023}The goal of this research is to more precisely examine and categorize the user attitudes. The BERT model is employed in this essay to categorize emotions.\cite{Selvakumar2022}The data included in the papers may then be specifically abstracted using this technique, followed by polarity-based categorization. Technology-assisted polarity allows the review or essay to be divided into good, negative, or neutral categories.\cite{Shaik2023} \cite{Peng2022}The emotion of the text, the topic, the entities, and the category of the phrase or sentence are all examined using natural language processing and other methods. Speech recognition, natural language production, and natural language understanding are difficulties that NLP must overcome.\cite{Hu2004}

The procedures for categorizing sensations are as follows:
\begin{itemize}
    \item data gathering
    \item data preparation
    \item Using BERT to classify emotions
\end{itemize}

\begin{figure}[ht]
    \centering
    \includegraphics[scale=0.5]{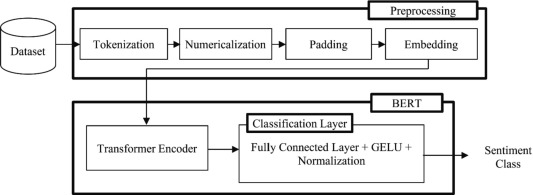}
    \caption{Proposed System Architecture}
    \label{fig:enter-label}
\end{figure}

\subsection{DATA COLLECTION}
\subsubsection{Data for training}
In this article, we will focus on the academic semester 2023-I. This semester's data set contains 87839 user suggestions regarding the teacher's teaching in the form of text, and these will be used to model the feelings classifier with the aid of BERT. The data set for the training comes from the Teacher Evaluations provided by the National University of the Altiplano each academic semester.

The reviews in both data sets are gathered in.csv file format and are provided as text documents.
\begin{table}[ht]
    \begin{tabular}{ p{1.2cm} p{7.2cm} }
         \hline
         Academic Department & Student response to teaching Performance\\
         \hline
         DERECHO& 	Tiene buena metodología de enseñanza\\
        TURISMO& 	Debe organizar su tiempo y planificar sus clases\\
        INGENIERIA CIVIL& 	Buen docente aunque considero que tiene mucho más conocimiento por impartir\\
        DERECHO& 	El docente es especialista en su área.\\
        INGENIERIA DE SISTEMAS	& El curso se ha desarrollado de manera VIRTUAL, sin embargo, considero que para un mayor aprendizaje y a la par de que haya una igualdad se debería realizar de manera PRESENCIAL.\\
        INGENIERIA QUIMICA	& El docente desarrolla las sesiones activamente y es especialista en su área.\\
         INGENIERIA CIVIL& Se desarrollan las clases de manera activa y el docente es especialista en su área.\\
        DERECHO	& Se desarrollan las clases de manera activa.\\
        INGENIRIA DE MINAS	& El docente es especialista en su área y desarrolla las clases de una manera amigable para el estudiante y a su vez profundizando los temas.\\
        \hline
    \end{tabular}
    \label{Tabla 1}
\end{table}

\subsubsection{Data for training}
The objective in this study is to be able to carry out a statistical analysis based on the feelings of the survey according to its polarity, the data in this study belong to the suggestions of the satisfaction survey carried out on teachers of a workshop course taught by the National University of the Altiplano.
\begin{table}[ht]
    \centering
    \begin{tabular}{ c l }
        \hline
         Level of satisfaction & Suggestions\\
         \hline
        Very satisfied & That the recorded sessions are permanently available to teachers\\
        Satisfied & Share the material with us\\
        little satisfied & Brevity and more interaction\\
        Very satisfied & Congratulations\\
        Satisfied & Facilitate PPTs\\
        Satisfied & Only punctuality has to be respected.\\
        Very satisfied & More courses because they are very interesting\\
        Satisfied & Specify the processes and responsible\\
        Satisfied & Deliver previous material\\
        Satisfied & Plan your activities for the whole year\\
        \hline
    \end{tabular}
    \label{Tabla 2}
\end{table}
The table shows 10 data, in general there is a total of 506 records of teachers

\subsection{DATA PROCESSING}

Records from the data set are given as input. The data is preprocessed using the following procedures for deep learning-based sentiment classification (BERT), in contrast to aspect-based sentiment classification.
\begin{itemize}
    \item \textit{Tokenization:} Creates a blob from reviews before turning it into a string of words.
    \item \textit{Numerization:} Each token in the corpus vocabulary must be assigned a distinct integer.
    \item \textit{Padding:} If a sentence is more than the prescribed length, omit the words; if it is shorter, add zeros.
    \item \textit{Embeddings:} To be used as model parameters, the words in the sentences are mapped to a 'n' dimensional vector.
\end{itemize}

\subsection{USING BERT TO CLASSIFY EMOTIONS}

BERT is utilized in this study to rate reviews according to sentiment. BERT receives the preprocessed data as input for sentiment categorization.\cite{Sun2017} The BERT model learns the correct connections between the complete string of words in a review using multiple self-attention based encoders with hidden layers. The bi-directional characteristic of BERT is employed to determine the viewpoint of a statement that is environment-focused.\cite{Shahade2023}

\subsection{TRANSFORMATION ENCODER}

The BERT transformer architecture is utilized in this work to do sentiment analysis on Spanish-language texts. Greater classification accuracy of feelings is achieved by fitting the pre-trained model to a Spanish corpus and supervised training using annotated data sets.\cite{Khaleghparast2023} \cite{Nur2023} The section provides information on how BERT was modified for this purpose, including the changes made to the output layer and the preprocessing methods that were used. This strategy marks a substantial advancement in Spanish natural language processing, offering up new opportunities for useful applications like opinion analysis and comprehending human language in this situation. The findings show promise in fields like emotion mining and the examination of opinions in Spanish-language texts.The pretrained bert model that was used is dccuchile/bert-base-spanish-wwm-uncased, in order to perform the sentiment analysis in Spanish \cite{CaneteCFP2020}

\begin{figure}[ht]
    \centering
    \includegraphics[scale=0.3]{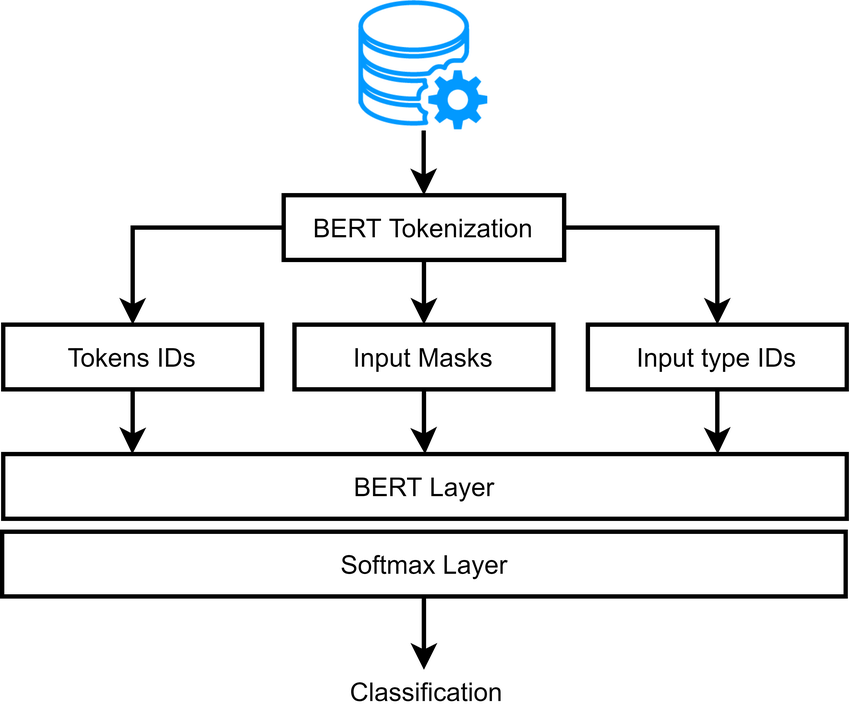}
    \caption{Architecture for Bert Encoder.}
    \label{fig:bert}
\end{figure}

\section{RESULTS}

The data set for this study's evaluation of teacher performance at the National University of the Altiplano includes 25,000 user comments for various teachers in various study programs, and 506 teacher suggestions from a workshop course were used to verify the model.

The training and test were conducted using the following ratios: 70:30, 80:20, and 90:10, in 10 ephocs having the following accuracys 93.5\% , 99.6\% and 94.3\% respectively. with the 80:20 ratio producing the best results of those. Initially, testing were conducted on the teacher evaluation data set itself. 

In addition, the proposed BERT model was compared to other models, and it was shown to perform better in classifying the presented data set's emotional content. When categorizing emotions from the teacher assessment data set, our system achieves an accuracy of about 99.6 \%.

Then, in order to better grasp the sentiments surrounding this course and to be able to improve in the upcoming events, this trained Bert model was used to categorize feelings in the workshop course that was made available to university teachers.

\subsection{Data Prediction}

Once we have the model already trained with our teacher evaluation data, we proceed to apply this model to our data from the workshop course provided to teachers, to later carry out a statistical analysis of the feelings (negative and positive) regarding this workshop course, likewise we show a cloud of words, in order to better understand and analyze the developed course.

\begin{figure}[ht]
    \centering
    \includegraphics[scale=0.35]{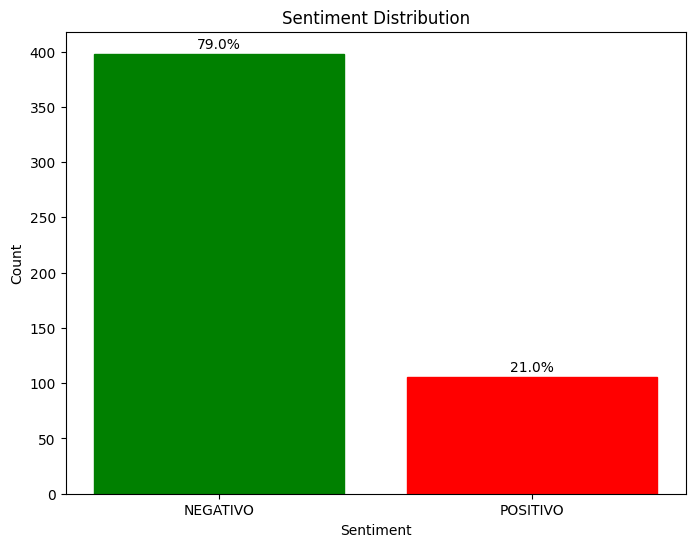}
    \caption{Distribution Sentiment for Prediction}
    \label{fig:distribution}
\end{figure}

\begin{figure}[ht]
    \centering
    \includegraphics[scale=0.25]{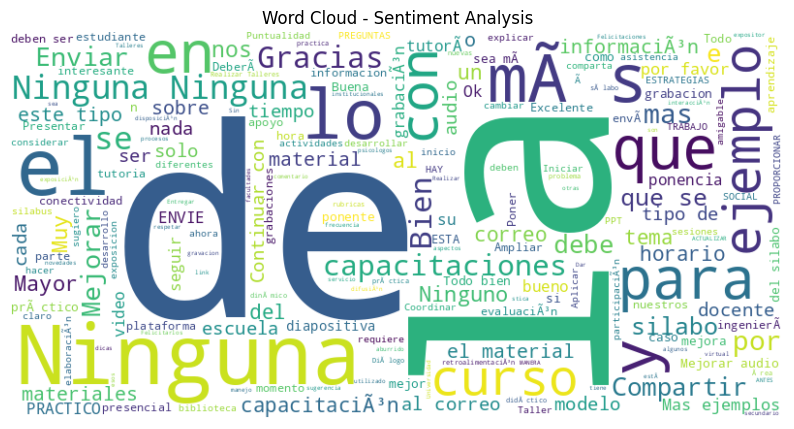}
    \caption{World cloud for Data}
    \label{fig:world}
\end{figure}

As can be seen, when carrying out the analysis of feelings for the workshop course offered to teachers, it is found that 79\% have a negative feeling regarding the developed course and we can see this in the word cloud Fig.\ref{fig:world}, texts how to improve, requesting slides and none that is being characterized as a negative feeling. and only 21 \% have a positive feeling regarding this course.

\section{DISCUSSION AND CONCLUSIONS}

Additionally, the effectiveness of BERT is evaluated in Fig. 2 in comparison to recent work on the same data set from \cite{Thet2010}, \cite{Kumar2019}, \cite{Maulana2020}, and \cite{Shaukat2020} for sentiment categorization. It demonstrates that BERT performs better than current literature-based efforts.

\begin{figure}[ht]
    \centering
    \includegraphics[scale=0.5]{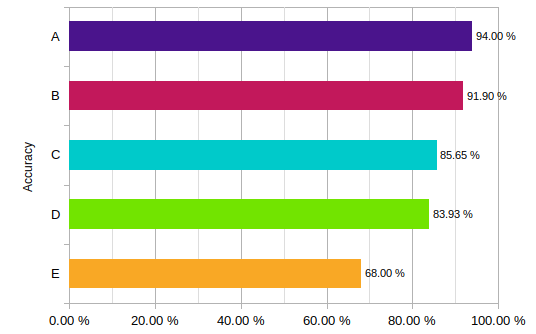}
    \caption{Accuracy comparison with existing works.}
    \label{fig:comparasion}
\end{figure}

\begin{enumerate}[label=(\Alph*)]
    \item Sentiment Analysis With BERT
    \item Lexicon with Multilayer Perceptron
    \item SVM with Information Gain
    \item Lexicon with Multilayer Perceptron
    \item Aspect Based Sentiment Analysis
\end{enumerate}

The suggested BERT-based sentimental classification outperforms existing ML and DL models and is able to accurately identify emotion polarity.\cite{Razali2021} Additionally taken into account when determining the emotion of reviews are word negations and intensifications. By experimenting with alternative algorithms for emotive analysis of user evaluations, this work might be further enhanced in the future. It is possible to shorten the additional processing time for a single review and improve the performance of the suggested design.



\newpage
\bibliographystyle{IEEEtran}
%
\bibliography{document}

\end{document}